\documentclass[11pt,a4paper]{article}
\usepackage[utf8]{inputenc}

\usepackage[usenames,dvipsnames,svgnames,table]{xcolor}
\usepackage{naaclhlt2018}
\usepackage{times}
\usepackage{latexsym}
\usepackage{amsmath}

\usepackage{comment}

\usepackage{CJK}
\newenvironment{zh}{\begin{CJK}{UTF8}{gbsn}}{\end{CJK}}
\newcommand{\textzh}[2]{\begin{zh}#1\end{zh}~\emph{#2}}
\usepackage{tipa}

\usepackage{booktabs}
\usepackage{makecell}

\usepackage{expex}
\lingset{everygla=,everyglb=\footnotesize}

\usepackage{graphicx}
\graphicspath{{images/}}
\usepackage[percent]{overpic}

\renewcommand{\|}{\mid}
\DeclareMathOperator*{\argmax}{arg\,max}
\newcommand{\word}[1]{\emph{#1}}

\newcommand{\colorPatch}[2][x]{
  \colorbox[HTML]{#2}{{\color[HTML]{#2}{\large #1}}}}

\usepackage{subcaption}

\newcommand{\figref}[1]{Figure~\ref{#1}}
\newcommand{\Figref}[1]{Figure~\ref{#1}}
\newcommand{\tabref}[1]{Table~\ref{#1}}

\newcommand{\secref}[1]{Section~\ref{#1}}

\aclfinalcopy %

\title{Generating Bilingual Pragmatic Color References}

\author{Will Monroe \\
  Computer Science Department \\
  Stanford University \\
  {\tt wmonroe4@cs.stanford.edu} \\\And
  Jennifer Hu \\
  Department of Mathematics \\
  Harvard University \\
  {\tt jenniferhu@college.harvard.edu} \\\AND 
  Andrew Jong \\
  Department of Computer Science \\
  San Jose State University \\
  {\tt andrewjong.cs@gmail.com} \\\And
  Christopher Potts \\
  Department of Linguistics \\
  Stanford University \\
  {\tt cgpotts@stanford.edu} \\}

\date{}

\begin{document}
\maketitle
\begin{abstract}
 Contextual influences on language often exhibit substantial cross-lingual
 regularities; for example, we are more verbose in situations that require finer
 distinctions. However, these regularities are sometimes obscured by semantic
 and syntactic differences. Using a newly-collected dataset of color reference games in Mandarin
 Chinese (which we release to the public), we confirm that a variety of constructions
 display the same sensitivity to contextual difficulty in Chinese and English. We
 then show that a neural speaker agent trained on bilingual data with a simple
 multitask learning approach displays more human-like patterns
 of context dependence and is more pragmatically informative than its monolingual
 Chinese counterpart. Moreover, this is not at the expense of language-specific
 semantic understanding: the resulting speaker model learns the different basic color
 term systems of English and Chinese (with noteworthy cross-lingual influences), and it
 can identify synonyms between the two languages using vector analogy operations on
 its output layer, despite having no exposure to parallel data.

\end{abstract}

\section{Introduction} \label{sec:intro}

In grounded communication tasks, speakers face pressures in choosing referential expressions that distinguish their targets from others in the context, leading to many kinds of pragmatic meaning enrichment. For example, the harder a target is to identify, the more the speaker will feel the need to refer implicitly and explicitly to alternatives to draw subtle contrasts \citep{Zipf49,Horn84,Levinson00}. However,
the ways in which these contrasts are expressed depend heavily on language-specific syntax and semantics.

\setlength{\fboxrule}{2pt}

\begin{figure}
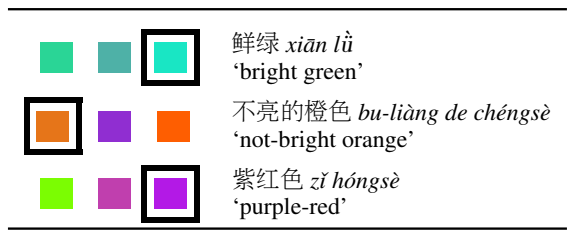

    \centering
    \footnotesize
    \renewcommand{\arraystretch}{2.5}
    \begin{tabular}{c@{\hskip 1ex}c@{\hskip 1ex}cl}
        \toprule
        \colorPatch{2AD596} & \colorPatch{4EB1A7} & \framebox{\negthickspace\colorPatch{19E6C4}} & 
            \makecell[cl]{\textzh{鲜绿}{xi\=an l\textipa{\`{\"u}}} \\ `bright green'} \\
        \framebox{\negthickspace\colorPatch{E67519}} & \colorPatch{902ed1} & \colorPatch{FE5E01} & 
            \makecell[cl]{\textzh{不亮的橙色}{bu-li\`ang de ch\'engs\`e} \\ `not-bright orange'} \\
        \colorPatch{7BFD02} & \colorPatch{C03FAE} & \framebox{\negthickspace\colorPatch{B319E6}} & 
            \makecell[cl]{\textzh{紫红色}{z\v{\i} h\'ongs\`e} \\ `purple-red'} \\
        \bottomrule 
    \end{tabular}
    \caption{Reference game contexts and utterances from our Chinese corpus. The boxed color is the target. Some color terms show differences between Chinese and English, such as \textzh{绿}{l\textipa{\`{\"u}}} `green' in the first example for a color that might be referred to with `blue' or `aqua' in English.}
    \label{fig:ref-game}
\end{figure}

In this paper, we seek to develop a model of contextual language production that captures language-specific syntax and semantics while also exhibiting responsiveness to contextual differences. We focus on a color reference game \citep{Rosenberg:Cohen:1964,Dale:Reiter:1995,Krahmer:vanDeemter:2012}
played in both English and Mandarin Chinese.
A reference game (\figref{fig:ref-game}) involves two agents, one designated the ``speaker'' and the other the ``listener''. The
speaker and listener are shown the same set of $k$ colors $C = \{c_1, \dotsc, c_k\}$ (in our experiments, 
$k = 3$), and one of these
colors $c_t$ is indicated secretly to the speaker as the ``target''. Both players share the same goal:
that the listener correctly guesses the target color. The speaker may communicate with the listener in
free-form natural-language dialogue to achieve this goal. Thus, a model of the speaker must process representations of the colors in the context and produce an utterance to distinguish the target color from the others. We evaluate a sequence-to-sequence speaker agent based on that of \citet{Monroe2017}, who also collected the English data we use; our Chinese data are new and were collected according to the same protocols.

While English and Chinese both use fairly similar syntax for color descriptions, our reference game is designed to elicit constructions that make reference to the context, and these constructions---particularly comparatives and negation---differ morpho-syntactically and pragmatically between the two languages. Additionally, Chinese is considered to have a smaller number of basic color terms \citep{BerlinKay1969}, which predicts markedness of more specific descriptions. %

Our primary goal is to examine the effects of \emph{bilingual} training: building one speaker trained on both English and Chinese data with a shared vocabulary, so that it can produce utterances in either language.  
The reference game setting offers an objective measure of success on the grounded
language task, namely, the speaker's ability to guide the listener to the target. We use this to address the tricky problem of speaker evaluation. Specifically, we use the speaker model and an application of Bayes' rule to infer the most likely target color given a human utterance, and we report the accuracy of that process at identifying the target color. We refer to this metric as \emph{pragmatic informativeness} because it requires not only accuracy but also effectiveness at meeting the players' shared goal \citep{Grice75}. 
A more formal definition and a discussion of alternatives are given in \secref{sec:metric}.

We show that a bilingually-trained model produces distributions over Chinese utterances that have higher pragmatic informativeness than a monolingual model.
An analysis of the learned word embeddings reveals that the bilingual model learns color synonyms between the two languages without being directly exposed to labeled pairs. However, using a context-independent color term elicitation task from \citet{BerlinKay1969} on our models, we show that the learned lexical meanings are largely faithful to each language's basic color system, with only minor cross-lingual influences. This suggests
that the improvements due to adding English data are not
primarily due to better representations of the input colors or lexical semantics alone.
The bilingual model does better resemble human patterns of utterance length
as a function of contextual difficulty, suggesting the
pragmatic level as one possible area of cross-lingual generalization.

\section{Data collection} \label{sec:data}

We adapted the open-source reference game framework of \citet{Hawkins15_RealTimeWebExperiments} to Chinese and followed the data collection protocols of \citet{Monroe2017} as closely as possible, in the hope that this can be the first step in a broader multilingual color reference project. We recruit pairs of players on Amazon Mechanical Turk in real time, randomly assigning one the role of the speaker and the other the listener. Players are self-reported Chinese speakers, but they must pass a series of Chinese comprehension questions in order to proceed, with instructions in a format preventing copy-and-paste translation. The speaker and listener are placed in a game environment in which they both see the three colors of the context and a chatbox. The speaker sends messages through the chatbox to describe the target to the listener, who then attempts to click on the target. This ends the round, and three new colors are generated for the next. Both players can send messages through the chatbox at any time. After filtering out extremely long messages (number of tokens greater than $4\sigma$ above the mean), spam games,\footnote{Some players found they could advance through rounds by sending duplicate messages. Games were considered spam if the game contained 25 or more duplicates.} and players who self-reported confusion about the game, we have a new corpus of 5,774 Chinese messages in color reference games, which we will release publicly. Data management information is given in Appendix~\ref{sec:irb}.

As in \citet{Monroe2017}, the contexts are divided into three groups of roughly equal size: in the \emph{far} condition (1,421 contexts), all the colors are at least a threshold distance $\theta$ from each other; in the \emph{split} condition (1,412 contexts), the target and one distractor are less than $\theta$ from each other, with the other distractor at least $\theta$ away from both; and in the \emph{close} condition (1,425 contexts), all colors are within $\theta$ from each other. We set $\theta = {}$20 by the CIEDE2000 color-difference formula \citep{Sharma2005}, with all colors different by at least 5.

\section{Human data analysis}

As we mentioned earlier, our main goal with this work is to investigate the effects of bilingual training on pragmatic language use. We first examine the similarities and differences in pragmatic behaviors between the English and Chinese corpora we use. The picture that emerges accords well with our expectations about pragmatics: the broad patterns are aligned across the two languages, with the observed differences mostly tracing to the details of their lexicons and constructions.

\subsection{Message length}

\begin{figure}[!t]
    \centering
    \includegraphics[width=0.8\columnwidth]{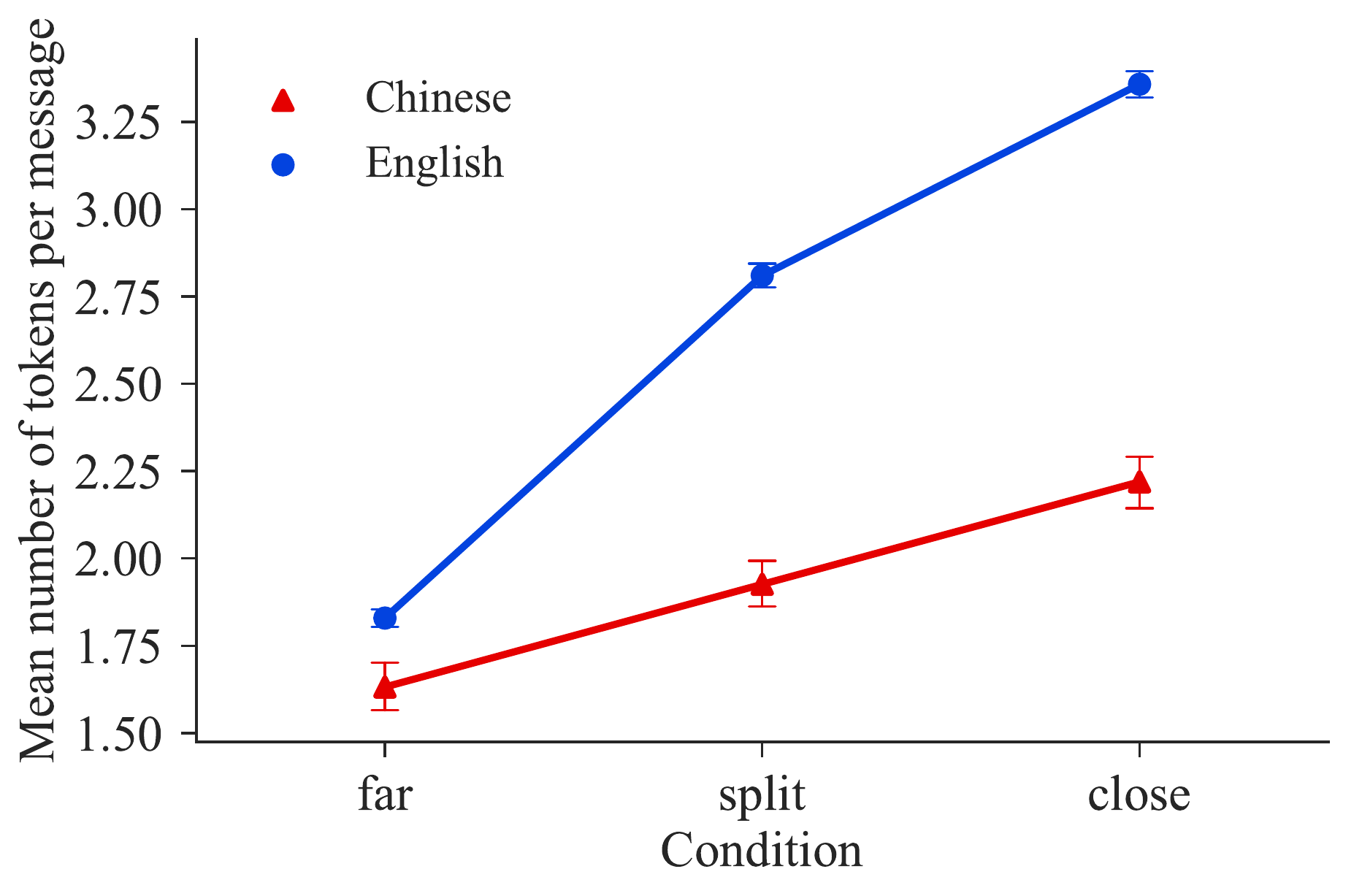}
\caption{Comparison of mean length of messages in English and Chinese. The \emph{split} and \emph{close} conditions have more similar context colors (\secref{sec:data}).}
\label{fig:length}
\end{figure}

We expect message length to correlate with the difficulty of the context: as the target becomes harder to distinguish from the distractors, the speaker will produce more complex messages, and length is a rough indicator of such complexity. %
To test this hypothesis, we used the Natural Language Toolkit (NLTK; \citealt{NLTKbook}) and Jieba \citep{Jieba:2015} to tokenize English and Chinese messages, respectively, and counted the number of tokens in both languages as a measure of message length. The results (\figref{fig:length}) confirm that in both languages, players become more verbose in more difficult conditions.\footnote{We do not believe that the overall drop in message length from English to Chinese reflects a fundamental difference between the languages; this has a few possible explanations, from Chinese messages taking the form of ``sentence segments'' \citep{WangQin2010} to differences in tokenization.}

\subsection{Specificity}

In the \emph{split} and \emph{far} conditions, the speaker must make fine-grained distinctions. A broad color term like \word{red} will not suffice if there are two reds, but more specific terms like \word{maroon} might identify the target. Thus, we expect specificity to increase as the difficulty of the context does. To assess this, we use WordNet \citep{WordNet98} to transform adjectives into derivationally-related noun forms, filter for nouns with \word{color} in their hypernym paths, and mark a message as ``specific'' if it contains at least one word with a hypernym depth greater than 7.

For Chinese, we translate to English via Google Translate, then measure the translated word using WordNet. It should be noted that this method has the drawback of obscuring differences between the two languages' color systems, as well as the potential for introducing noise due to errors in automatic translation. Though Mandarin variations of WordNet exist, we chose this translation method to standardize hypernym paths for both languages. Differences in ontology decisions between lexical resources prevent straightforward cross-lingual comparisons of hypernym depths, while automatic translation to a common language ensures the resulting hypernym paths are directly comparable.

\begin{figure}[!t]
    \centering
    \includegraphics[width=0.8\columnwidth]{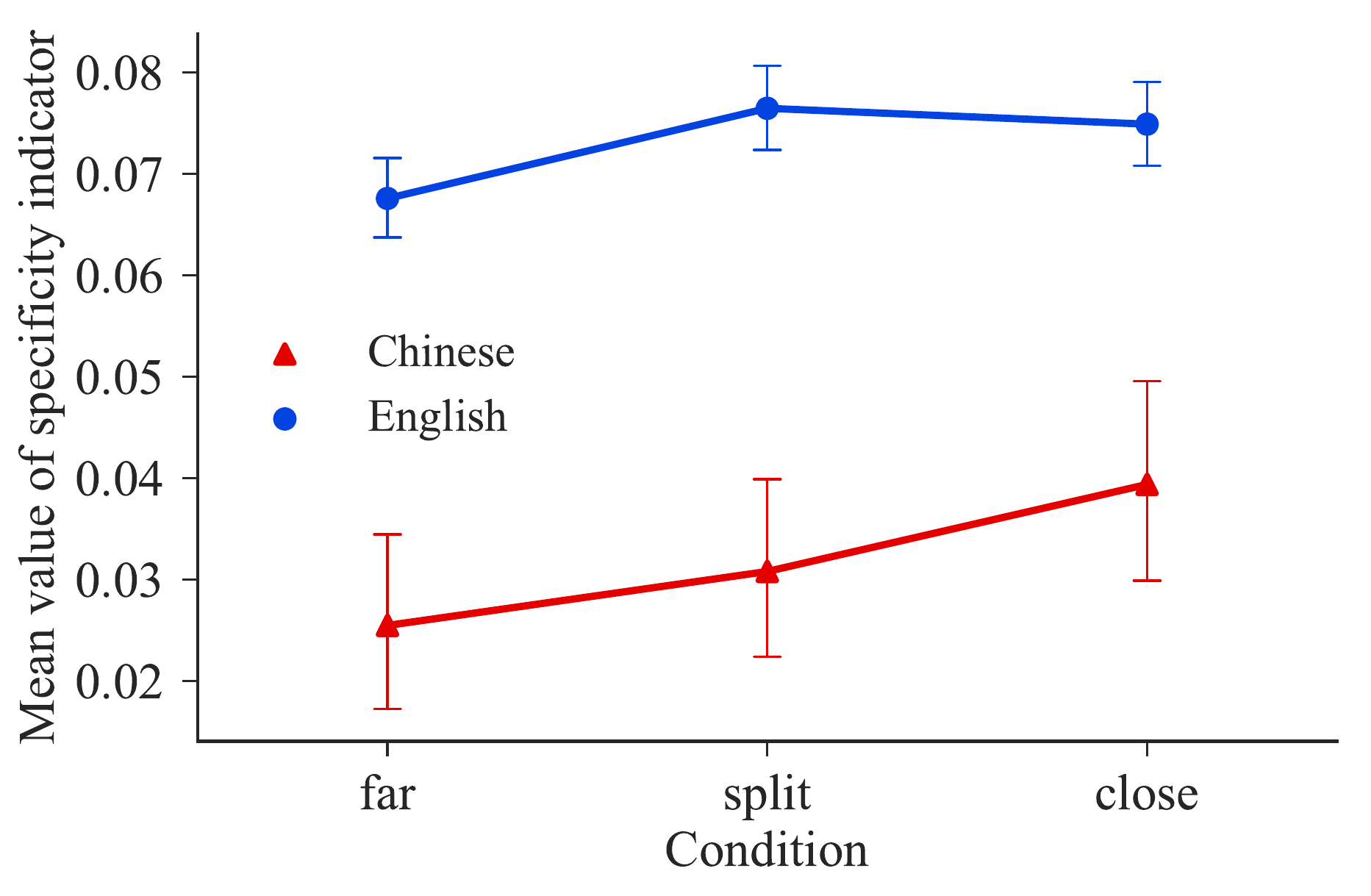}
    \caption{Comparison of WordNet specificity in Chinese and English.}
    \label{fig:specificity}
\end{figure}

\Figref{fig:specificity} summarizes the results of this measurement. In general, the usage of high-specificity color words increases in more difficult conditions, as expected. However, we see that Chinese speakers use them significantly less than English speakers. Instead, Chinese speakers use nominal modifiers, such as \textzh{草}{c\v{a}o} `grass' and \textzh{海}{h\v{a}i} `ocean', which do not contain ``color'' in their hypernym paths and are thus not marked as high-specificity. 
To quantify this observation, we annotated random samples of 200 messages from each language for whether they contained nominal color descriptions, and found that 3.5\% of the English messages contain such nominals versus 13.5\% of the Chinese messages.

The use of nominal modifiers as opposed to adjectives (`dark orange', `dull brown') is arguably expected given the claims of \citet{BerlinKay1969} and others that Chinese has fewer basic color terms than English, thus requiring more visually evocative modifiers to clarify distinctions between similar hues. (This isn't a complete explanation, since Chinese is rich in narrow but rare non-basic color terms. For the cases where Chinese has an appropriate narrow color term, it is possible that speakers make a pragmatic decision to avoid obscure vocabulary in favor of more familiar nouns.)

\subsection{Comparatives, superlatives, and negation}

To detect comparative and superlative adjectives in English, we use NLTK POS-tagging, which outputs JJR and RBR for comparatives, and JJS and RBS for superlatives. In Chinese, we look for the tokens \textzh{更}{g\`eng} `more' and \textzh{比}{b\v{\i}} `comparatively' to detect comparatives and \textzh{最}{zu\`{\i}} `most' to detect superlatives. We detect negation by tokenizing messages with NLTK and Jieba and then looking for the tokens \emph{not} and \emph{n't} in English and corresponding \textzh{不}{b\`u} 
and \textzh{没}{m\'ei} 
in Chinese.

\begin{figure}[!t]
    \centering
    \begin{subfigure}[b]{\columnwidth}
    \centering
    \includegraphics[width=0.8\textwidth]{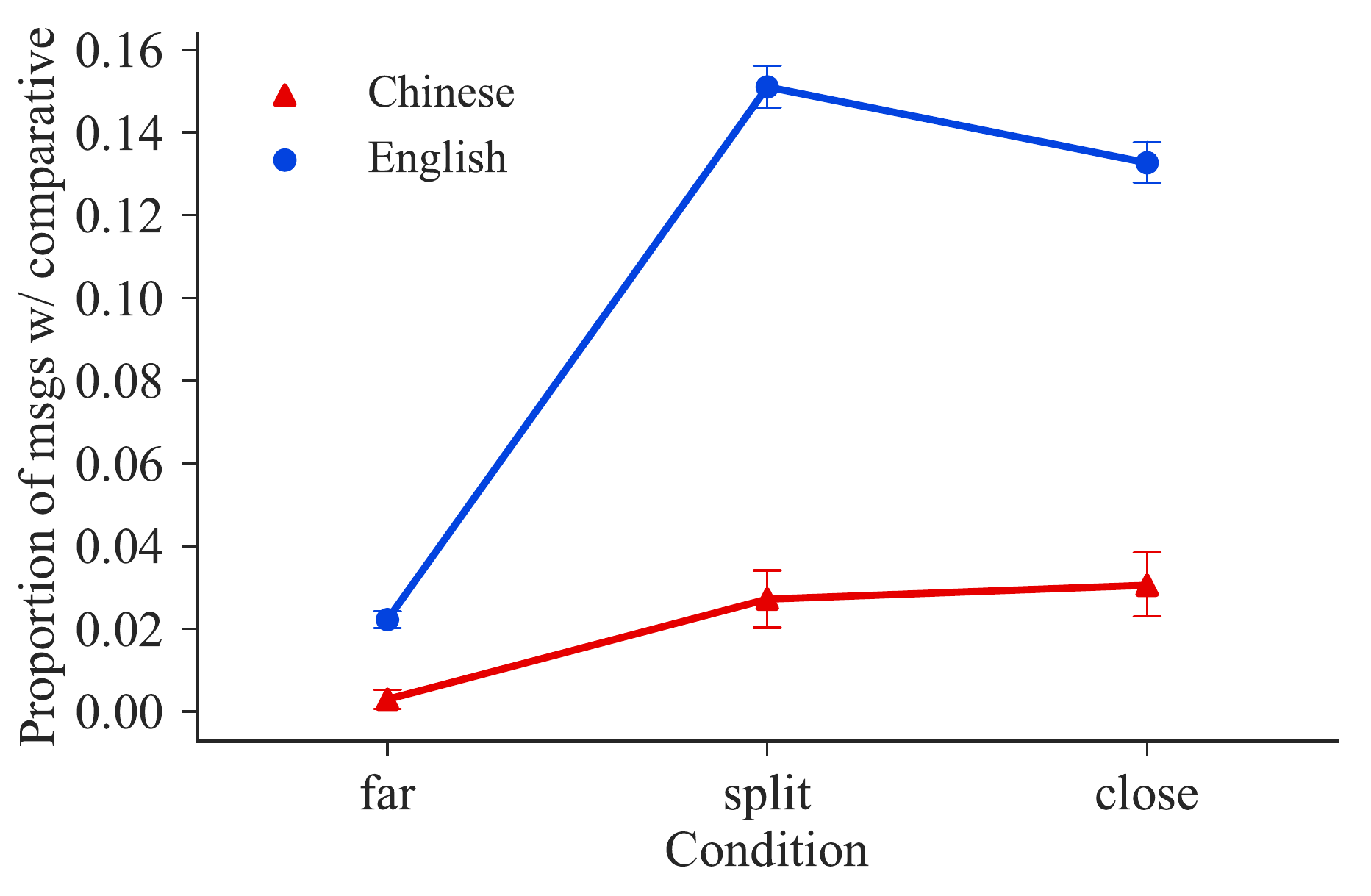}
    \caption{Usage of comparative adjectives in Chinese and English.}
    \label{fig:comparative}
    \end{subfigure}

    \begin{subfigure}[b]{\columnwidth}
    \centering
    \includegraphics[width=0.8\textwidth]{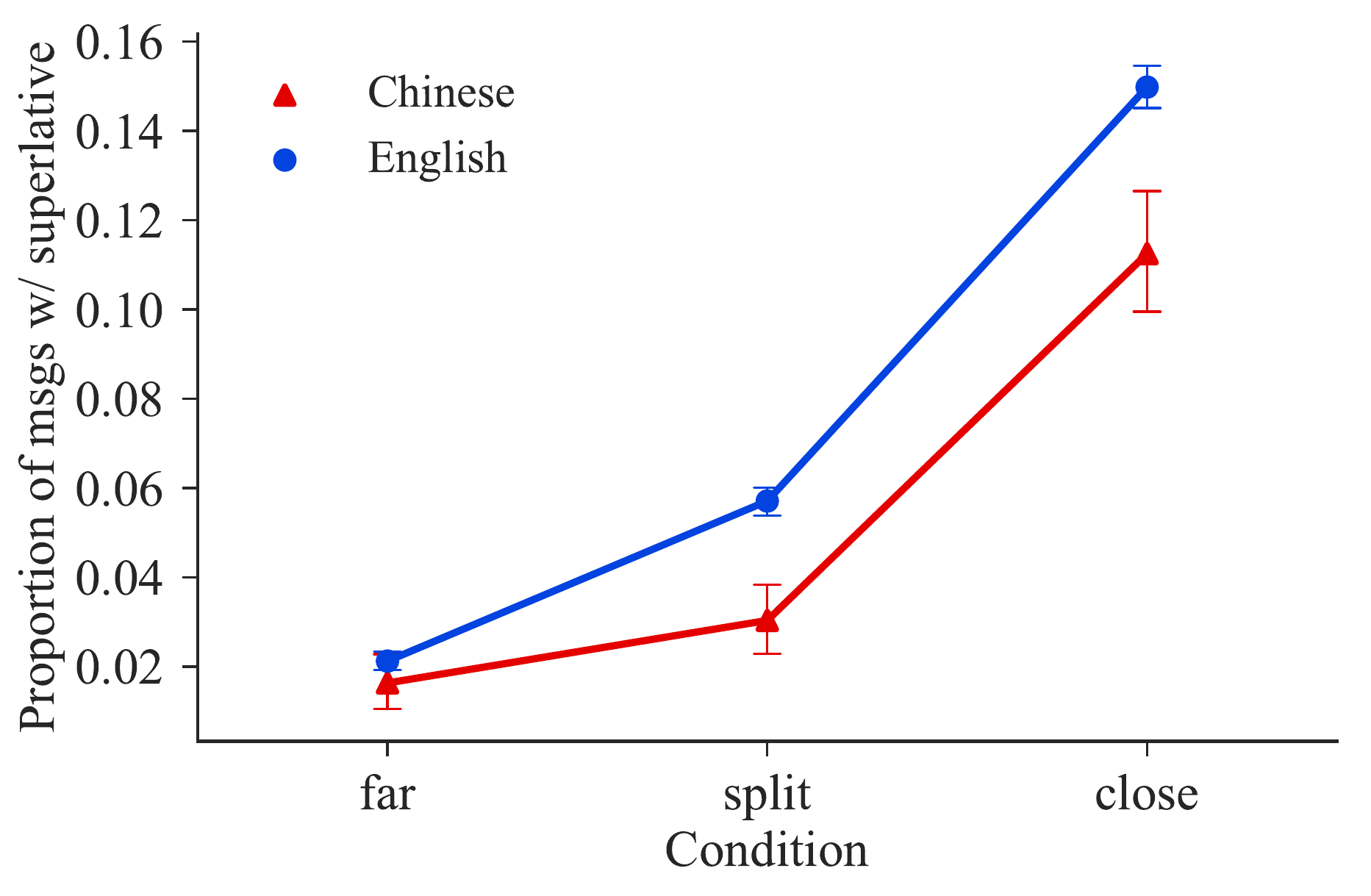}
    \caption{Usage of superlative adjectives in Chinese and English.}
    \label{fig:superlative}
    \end{subfigure}
    
    \begin{subfigure}[b]{\columnwidth}
    \centering
    \includegraphics[width=0.8\textwidth]{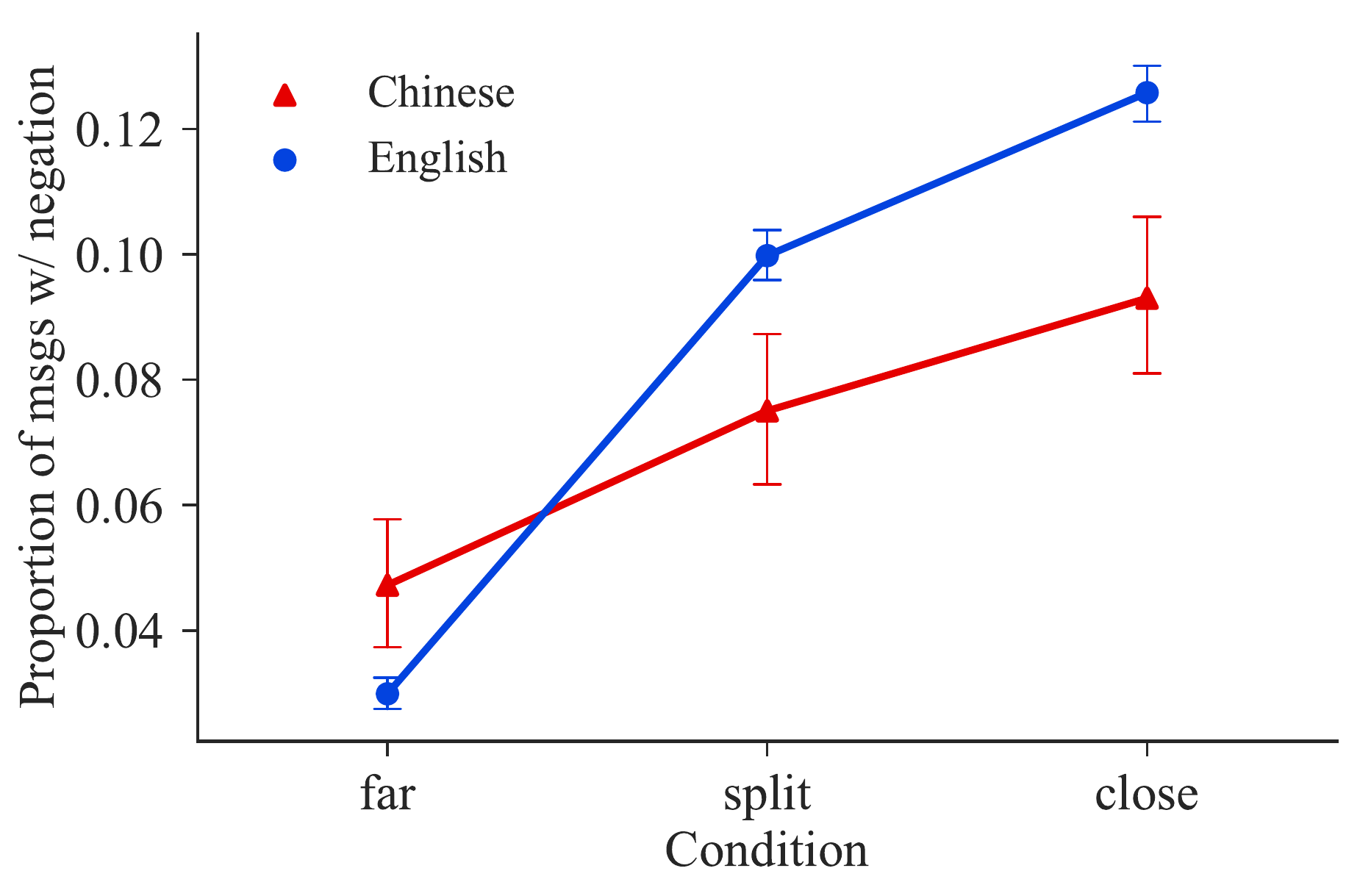}
    \caption{Usage of negation in Chinese and English.}
    \label{fig:negation}
    \end{subfigure}
\caption{Comparison of usage of comparatives, superlatives, and negation in English and Chinese.}
\label{fig:comp-super-neg}
\end{figure}

These statistics are shown in \figref{fig:comp-super-neg}. Both languages exhibit similar trends for superlative adjectives. In English, comparatives are used most frequently in the \emph{split} condition and second most frequently in the \emph{close} condition, while in Chinese, they occur at around the same rate in the \emph{split} and \emph{close} conditions. The literature is not conclusive about the source of these differences. \citet{Xia:2014} argues that complex attributives are rarely used and sound ``syntactically deviant or Europeanized'' \citep{Zhu:1982,Xie:2001} in Chinese, citing the left-branching nature of the language as restricting attributives in length and complexity. There are also conflicting theories on the markedness of gradable adjectives in Chinese \citep{Grano:2012,Ito:2008}; such markedness may contribute to the frequency at which comparative forms are used.

We also see that both languages follow the same general trend of using negation more frequently as the condition becomes more difficult.

\section{Models}\label{sec:models}

We build and evaluate three artificial agents on this reference game task, two trained
on monolingual descriptions (one for each language) and one on bilingual descriptions.
We base these models on the basic speaker architecture
from \citet{Monroe2017}. The monolingual speakers 
represent the context by passing all the
context colors as input to a long short-term memory (LSTM) sequence encoder, then concatenating
this representation with a word vector for
each previous output token as the input to an LSTM decoder that produces a color description token-by-token. This defines a distribution over descriptions $u$ conditioned on the target and context, $S(u \| c_t, C)$.

To accommodate bilingual training with this architecture, we expand the
vocabulary to include English and Chinese words, and we add a flag $\ell$ to the input specifying whether the model's
output should be in English ($\ell = 0$) or Chinese ($\ell = 1$):
\[S(u \| \ell, c_t, C) = \prod_{i=1}^{|u|} s(u_i \| u_{1..i-1}, \ell, c_t, C)\]
The flag $\ell$ is embedded as a single additional dimension that is concatenated alongside the context and input (previous token)
vectors for the encoder. See Appendix~\ref{sec:appendix} for additional training details.

\subsection{Pragmatic informativeness} \label{sec:metric}

As mentioned in \secref{sec:intro}, we evaluate the two models on a measure of \emph{pragmatic informativeness}: how well does
the model represent a human speaker, such that a generative model of a listener can be built from it to interpret utterances?
Formally, for a speaker $S(u \| \ell, c_t, C)$ and an example consisting of an utterance, language identifier, and
color context $(u, \ell, C)$, we identify the $t^*$ that maximizes the probability of $u$ according to $S$:
\[t^* = \argmax_t S(u \| c_t, C)\]
That is, $L$ uses a noisy-channel model with a uniform prior over target colors and $S$ as a generation model to infer the most likely target color given the input utterance. The pragmatic informativeness of a speaker is the proportion of target colors in a test set correctly identified by $t^*$.

One drawback of this metric is it does not evaluate how faithful the model is to the overall distribution of human utterances, only the relative conditional likelihoods of human utterances for different target colors. In practice, since the agents are trained to minimize log likelihood, we do not observe our agents frequently producing wildly unhumanlike utterances; however, this is a caveat to keep in mind for evaluating agents that do not naturally approximate a language model.

The understanding model implied in this metric is equivalent to a version of the Rational Speech Acts model of pragmatic language understanding \cite{Frank2012,GoodmanFrank16_RSATiCS}, or the \emph{pragmatic posterior} of the
Rational Observer model \cite{McMahan:Stone:2015}. An important difference between our 
speaker model and those in the work cited above is that our speaker model is a neural 
network that makes a combined judgment of applicability (semantic appropriateness) and 
availability (utterance prior), instead of modeling the two components separately.
However, we stop short of directly predicting the referent of an expression
discriminatively, as is done by e.g.\ \citet{Kennington2015}, so as to require a
model that is usable as a speaker.

A related metric is \emph{communicative success} as defined by \citet{Golland2010}, which judges the speaker by the accuracy of a human listener when given model-produced utterances. Our pragmatic informativeness metric instead gives a model-derived listener human utterances and assesses its accuracy at identifying colors. Pragmatic informativeness has the advantage of not requiring additional expensive human labeling in response to model outputs; it can be assessed on an existing collection of human utterances, and can therefore be considered an automatic metric.

\subsection{A note on perplexity}

Perplexity is a common intrinsic evaluation metric for generation
models.\footnote{Two other intrinsic metrics, word error
rate (WER) and BLEU \cite{Papineni2002}, were at or worse than
chance despite qualitatively adequate speaker outputs, due to high diversity in valid 
outputs for similar contexts. This problem is common in dialogue tasks,
for which BLEU is known to be an ineffective speaker evaluation metric \cite{Liu2016}.}
However, for comparing monolingual and bilingual models,
we found perplexity to be
unhelpful, owing largely to its vocabulary-dependent definition. Specifically, if we fix the vocabulary in advance to include tokens
from both languages, then the monolingual model performs
unreasonably poorly, and bilingual training helps immensely. However, this is an unfair comparison: the monolingual model's high perplexity
is dominated by low probabilities assigned to rare tokens in the opposite-language data that it did not see. Thus, perplexity ceases to
be a measure of language modeling ability and assumes the role of a proxy for the out-of-vocabulary rate.

On the other hand, if we define the output vocabulary
to be the set of tokens seen at least $n$ times in training ($n={}$1 and 2 are common), then monolingual training yields
better perplexity than bilingual training, but mainly because including opposite-language training data forces the
bilingual model to predict more rare words that would otherwise be replaced with $\langle$unk$\rangle$.\footnote{The rare
words that make this difference are primarily the small number of English words that were used by the Chinese-language participants;
no Chinese words were observed in the English data from \citet{Monroe2017}} %
This produces the counterintuitive result that perplexity initially goes \emph{up} (gets worse) when increasing the amount of
training data. (As a pathological case, with no training data, a model can get a perfect perplexity
of 1 by predicting $\langle$unk$\rangle$ for every token.)

\section{Experimental results and analysis}

\begin{table}[t]
\centering
\begin{tabular}{llcc}
\toprule
test & train & dev acc & test acc \\
\midrule
en     & en    & \textbf{80.51} & \textbf{83.06} \\
       & en+zh & 79.73          & 81.43 \\ %
\midrule
zh     & zh    & 67.16          & 67.75 \\
       & en+zh & \textbf{71.81} & \textbf{72.89} \\ %
\bottomrule
\end{tabular}
\caption{Pragmatic informativeness scores (\%) for monolingual and bilingual speakers.}
\label{tab:accuracy}
\end{table}

Pragmatic informativeness of the models on English and Chinese data is shown in \tabref{tab:accuracy}. The main result is that training a
bilingual model helps compared to a Chinese monolingual one; however, the benefit is asymmetrical, as training on monolingual English data is superior for English data to training on a mix of
Chinese and English. All differences in \tabref{tab:accuracy} are significant at $p < {}$0.001 \citep[approximate permutation test, 10,000 samples;][]{Pado2006}, except for the
decrease on the English dev set, which is significant at $p < {}$0.05.

\begin{figure}[t]
    \centering
    \begin{subfigure}[b]{0.465\columnwidth}
    \centering
    \includegraphics[width=\textwidth]{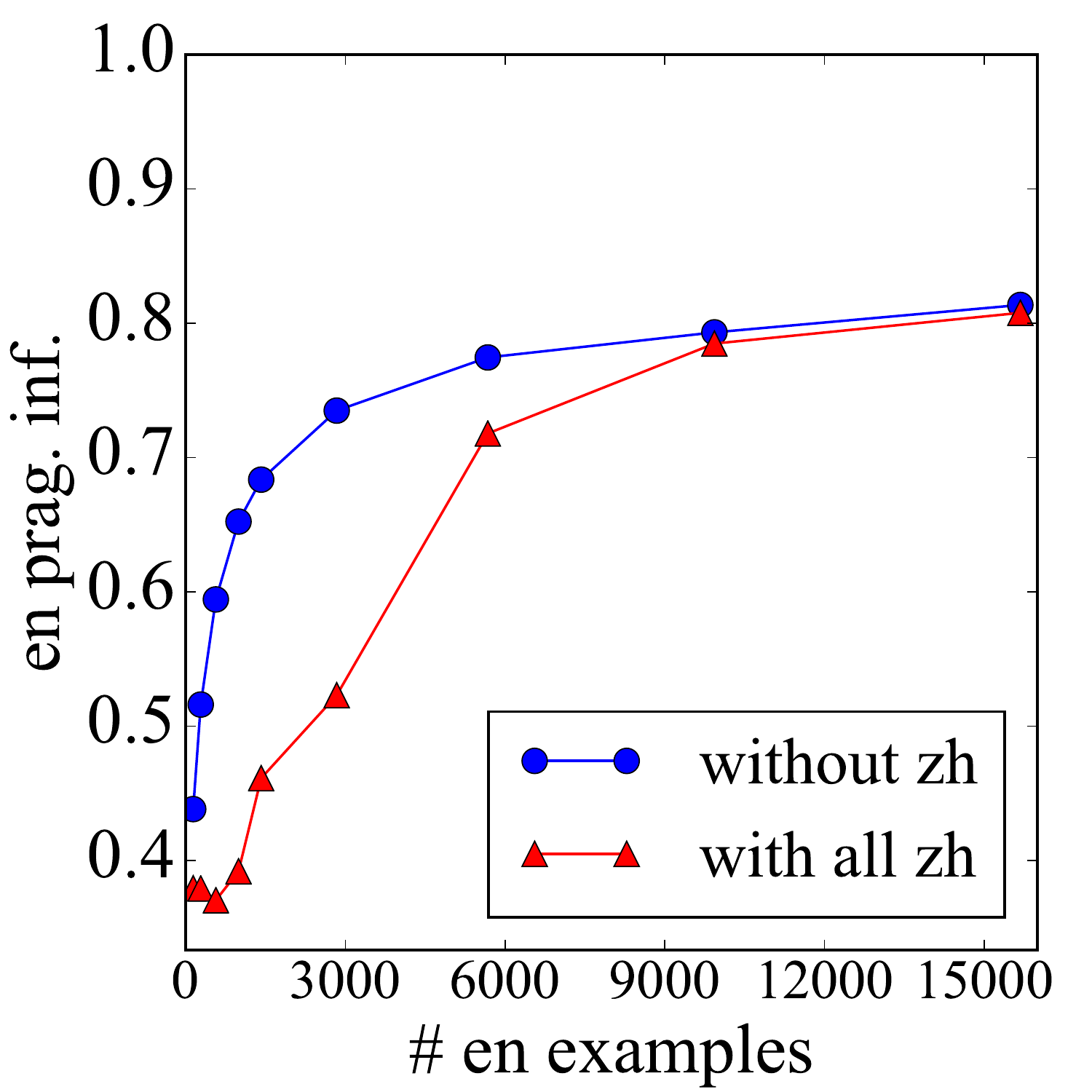}
    \caption{English without any / with all Chinese}
    \label{fig:lowdata-azh2en}
    \end{subfigure}
    \hspace{0.04\columnwidth}
    \begin{subfigure}[b]{0.465\columnwidth}
    \centering
    \includegraphics[width=\textwidth]{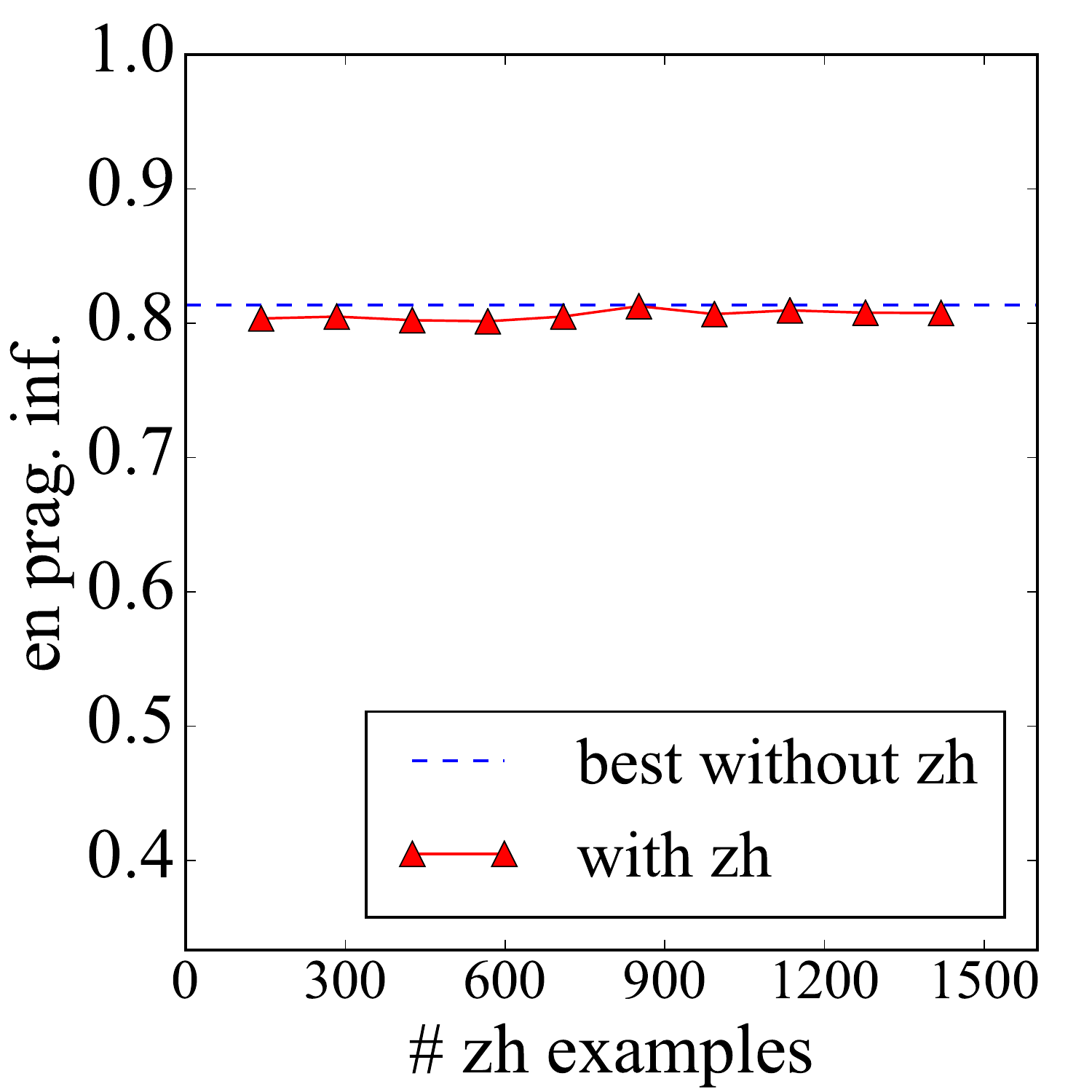}
    \caption{All English, varying amount of Chinese data}
    \label{fig:lowdata-zh2aen}
    \end{subfigure}

    \begin{subfigure}[b]{0.465\columnwidth}
    \centering
    \includegraphics[width=\textwidth]{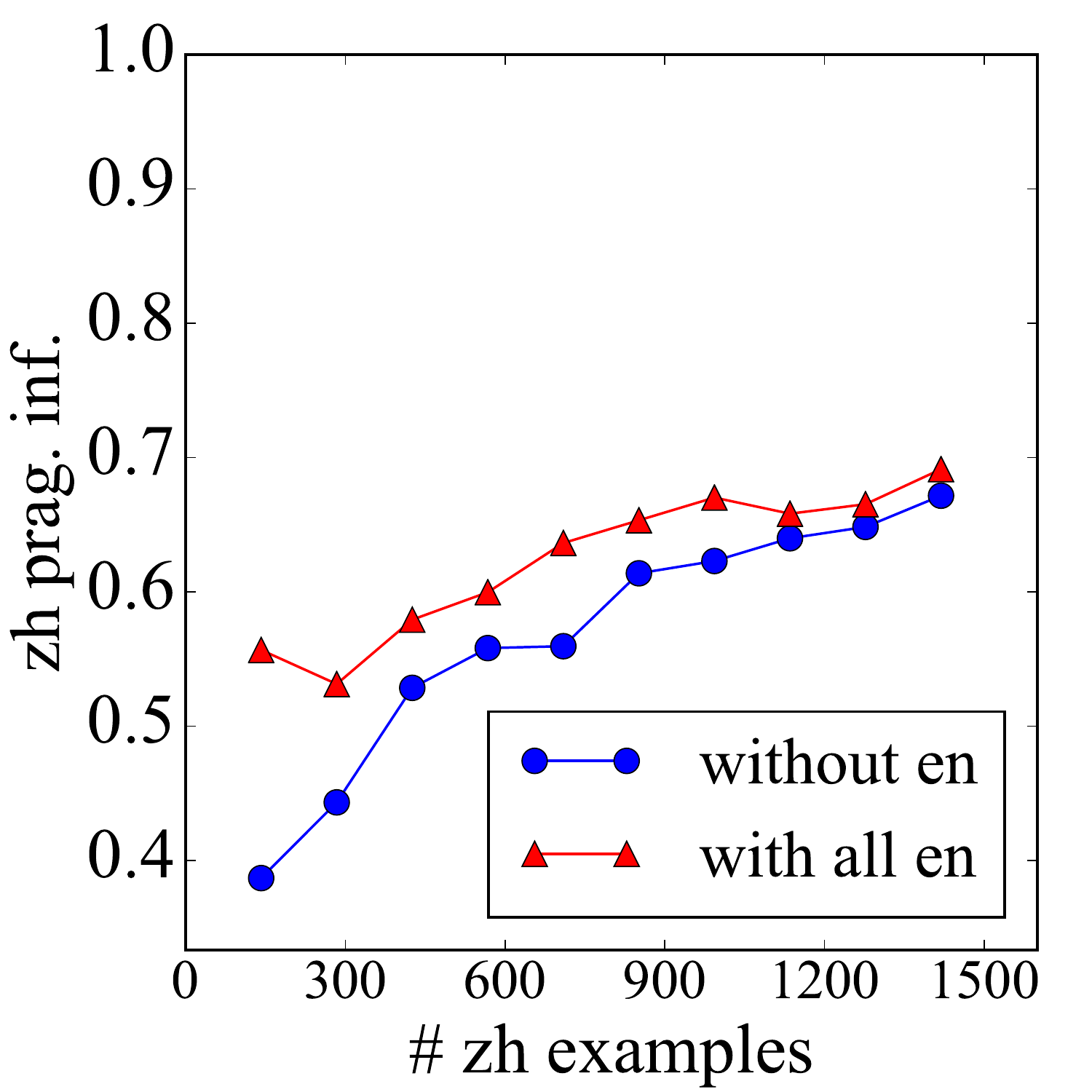}
    \caption{Chinese without any / with all English}
    \label{fig:lowdata-aen2zh}
    \end{subfigure}
    \hspace{0.04\columnwidth}
    \begin{subfigure}[b]{0.465\columnwidth}
    \centering
    \includegraphics[width=\textwidth]{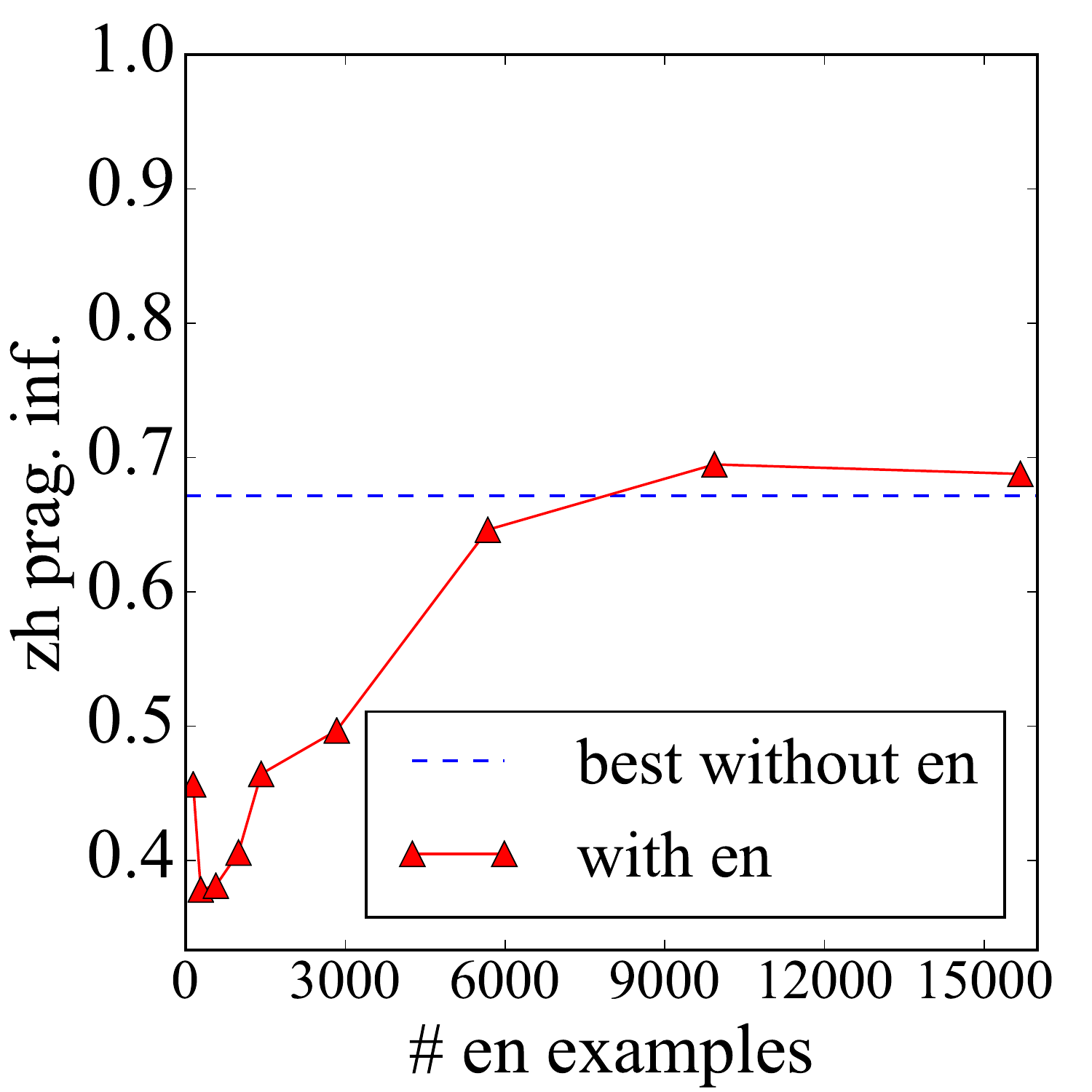}
    \caption{All Chinese, varying amount of English data}
    \label{fig:lowdata-en2azh}
    \end{subfigure}
\caption{Pragmatic informativeness (dev set) for different amounts and languages of training data.}
\label{fig:lowdata}
\end{figure}

An important difference between our corpora is that the English dataset is an order of magnitude larger than the Chinese. Intuitively, we expect adding more training data on the same task will
improve the model, regardless of language. However, we find that the effect of dataset size is not so straightforward. In fact, the differences in training set size convey a non-linear benefit.
\Figref{fig:lowdata} shows the pragmatic informativeness of the monolingual and bilingual speakers on the development set
as a function of dataset size (number of English and Chinese utterances). The blue curves (circles) in the plots on the left, 
\figref{fig:lowdata-azh2en} and \figref{fig:lowdata-aen2zh}, are standard learning curves for the monolingual 
models, and their parallel red curves (triangles) show the pragmatic informativeness of the bilingual model with the
same amount of in-language data plus all available data in the opposite language. The plots on the right, 
\figref{fig:lowdata-zh2aen} and \figref{fig:lowdata-en2azh}, show the effect of gradually adding 
opposite-language data to the bilingual model starting with all of the in-language data.

Overall, we see that adding all English data consistently helps the Chinese monolingual model, whereas
adding all Chinese data consistently hurts the English monolingual model (though with diminishing effects
as the amount of English data increases). Adding small amounts of English data---especially amounts comparable
to the size of the Chinese dataset---\emph{decreases} accuracy of the Chinese model dramatically. This suggests an
interaction between the total amount of data and the effect of bilingual training: a model trained on
a moderately small number of in-language examples can benefit from a much larger training set in another
language, but combining data in two languages is detrimental when both datasets are very small and has
very little effect when the in-language training set is large. This implies a benefit primarily in low-resource
settings, which agrees with the findings of \citet{Johnson2016} using a similar architecture for machine translation.

\subsection{Bilingual lexicon induction} \label{sec:lexicon-induction}

\begin{table}[t]
\centering
\footnotesize
\begin{tabular}{llll}
\toprule
\textbf{zh} & \textbf{en} & \textbf{en} & \textbf{zh} \\
\midrule
\textzh{绿色}{} `green' & \textbf{green}   & green & \textbf{\textzh{绿}{} `green'} \\
\textzh{紫色}{} `purple' & \textbf{purple} & blue & \textbf{\textzh{蓝}{} `blue'} \\
\textzh{蓝色}{} `blue' & purple            & purple & \textzh{蓝}{} `blue' \\
\textzh{灰色}{} `grey' & \textbf{grey}     & bright & \textbf{\textzh{鲜艳}{} `bright'} \\
\textzh{亮}{} `bright' & \textbf{bright}   & pink & \textbf{\textzh{粉色}{} `pink'} \\
\textzh{灰}{} `grey' & -er                 & grey & \textbf{\textzh{灰}{} `grey'} \\
\textzh{蓝}{} `blue' & \textit{teal}       & dark & \textbf{\textzh{暗}{} `dark'} \\
\textzh{绿}{} `green' & \textbf{green}     & gray & \textbf{\textzh{灰}{} `grey'} \\
\textzh{紫}{} `purple' & \textbf{purple}   & yellow & \textbf{\textzh{黄色}{} `yellow'} \\
\textzh{草}{} `grass' & \textit{green}     & light & \textzh{最}{} `most' \\
\bottomrule
\end{tabular}
\caption{Bilingual lexicon induction from Chinese to English (first two columns) and vice versa
(last two). Correct translations in \textbf{bold}, semantically close words in \textit{italic}.}
\label{tab:lexicon}
\end{table}

To get a better understanding of the influence of the bilingual training on the model's lexical
representations in the two languages, we extracted the weights of the final softmax layer of the bilingual
speaker model and used them to induce a bilingual lexicon with a word vector analogy task.
For two pairs of lexical translations, \textzh{蓝色}{l\'ans\`e} $\rightarrow$ ``blue'' and ``red''
$\rightarrow$ \textzh{红}{h\'ong}, we took the difference between the source language word vector and
the target language word vector. To ``translate'' a word, we added this ``translation vector'' to the
word vector for the source word, and found the
word in the opposite language with the largest inner product to the resulting vector.
The results are presented in \tabref{tab:lexicon}. We identified the 10 most frequent color-related words in each language to translate. (In other words,
we did not use this process to find translations of function words like ``the'' or the Chinese
nominalization/genitive particle \textzh{的}{de}, but we show proposed translations that were not
color-related, such as \textzh{灰}{hu\={\i}} being translated as the English comparative ending ``-er''.)
The majority of common color words are translated correctly by this simple method, showing that the vectors in
the softmax layer do express a linear correspondence between the representation of synonyms in the two languages.

\subsection{Color term semantics} \label{sec:wcs}

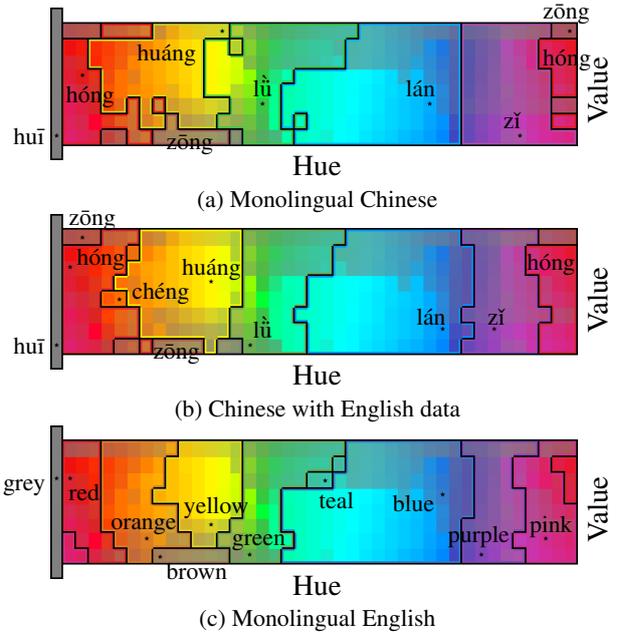
\begin{figure}[t]
    \centering

    \begin{subfigure}[b]{\columnwidth}
    \centering    
    \begin{overpic}[width=0.9\textwidth]{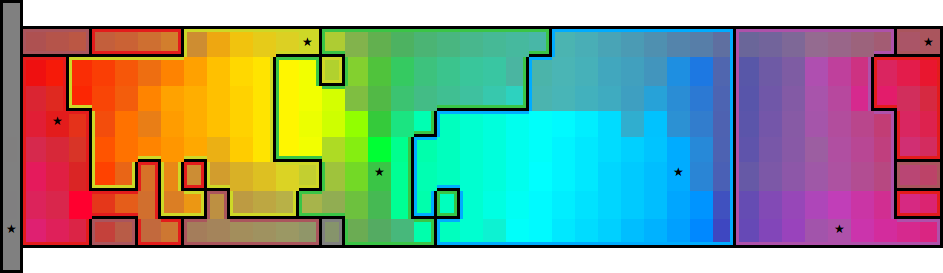}
        \put (-7, 3) {\footnotesize{hu\={\i}}}
        \put (3, 11) {\footnotesize{h\'ong}}
        \put (22, 2) {\footnotesize{z\=ong}}
        \put (16.5, 19) {\footnotesize{hu\'ang}}
        \put (38.5, 12) {\footnotesize{l\textipa{\`{\"u}}}}
        \put (67.5, 12) {\footnotesize{l\'an}}
        \put (86, 6) {\footnotesize{z\v{\i}}}
        \put (93.5, 18) {\footnotesize{h\'ong}}
        \put (93.5, 27) {\footnotesize{z\=ong}}

        \put (46, -3) {Hue}
        \put (102, 7) {\rotatebox{90}{Value}}
    \end{overpic}
    \vspace*{1ex}
    \caption{Monolingual Chinese}
    \label{fig:wcs-front-zh-mono}
    \end{subfigure}

    \begin{subfigure}[b]{\columnwidth}
    \centering
    \begin{overpic}[width=0.9\textwidth]{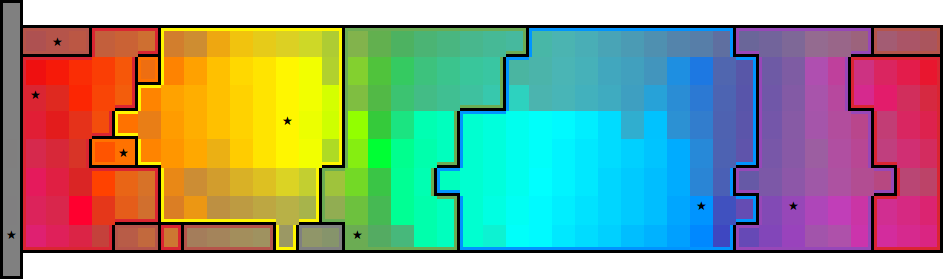}
        \put (-7, 3) {\footnotesize{hu\={\i}}}
        \put (3.5, 27.5) {\footnotesize{z\=ong}}
        \put (5, 20) {\footnotesize{h\'ong}}
        \put (15.5, 13.5) {\footnotesize{ch\'eng}}
        \put (19.5, 2) {\footnotesize{z\=ong}}
        \put (25, 18) {\footnotesize{hu\'ang}}
        \put (38.5, 6) {\footnotesize{l\textipa{\`{\"u}}}}
        \put (69.5, 8.5) {\footnotesize{l\'an}}
        \put (83, 8.5) {\footnotesize{z\v{\i}}}
        \put (90.5, 19) {\footnotesize{h\'ong}}

        \put (46, -3) {Hue}
        \put (102, 7) {\rotatebox{90}{Value}}
    \end{overpic}
    \vspace*{1ex}
    \caption{Chinese with English data}
    \label{fig:wcs-front-zh-bi}
    \end{subfigure}

    \begin{subfigure}[b]{\columnwidth}
    \centering
    \begin{overpic}[width=0.9\textwidth]{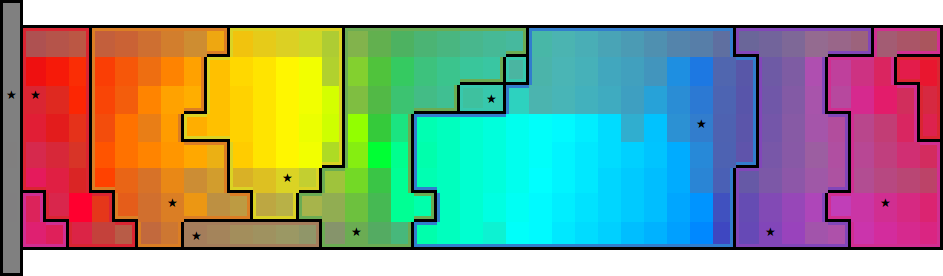} %
        \put (-9, 17) {\footnotesize{grey}}
        \put (3.5, 15) {\footnotesize{red}}
        \put (11.5, 9.5) {\footnotesize{orange}}
        \put (22, 0) {\footnotesize{brown}}
        \put (25.3, 12.5) {\footnotesize{yellow}}
        \put (34.5, 6.5) {\footnotesize{green}}
        \put (51, 13.5) {\footnotesize{teal}}
        \put (65, 13) {\footnotesize{blue}}
        \put (75.5, 7) {\footnotesize{purple}}
        \put (91, 9) {\footnotesize{pink}}

        \put (46, -3) {Hue}
        \put (102, 7) {\rotatebox{90}{Value}}
    \end{overpic}
    \vspace*{1ex}
    \caption{Monolingual English}
    \label{fig:wcs-en-mono}
    \end{subfigure}

\caption{Color term lexica: colors in the World Color Survey
palette grouped by highest-probability description, averaged over 10 randomly-generated pairs
of distractor colors.
The color that results in the highest probability of each description is marked with a star. English influences on the bilingual model include the appearance
of \textzh{橙色}{ch\'engs\`e} `orange' and narrowing of \textzh{黄色}{hu\'angs\`e} `yellow' and \textzh{绿色}{l\textipa{\`{\"u}}s\`e} `green'.} 
\label{fig:wcs}
\end{figure}

The above experiment suggests that the bilingual model has learned word semantics in ways that discover
translation pairs. However, we wish to know whether bilingual training has resulted in changes to the
model's output distribution reflecting differences in the two languages' color systems. To evaluate this,
we performed an experiment similar to the basic color term elicitations
in the World Color Survey \citep[WCS;][]{BerlinKay1969} on our models. For each of the 330 colors in the 
original WCS, we presented that color to our monolingual and bilingual models and recorded the
most likely color description according to the conditional language model. Our models require a three-color
context to produce a description; as an approximation to eliciting context-insensitive color terms, we gave the
model ten contexts with randomly generated (uniform in H, S, and V) distractor colors and averaged the language
model probabilities. We also identified, for each color term produced as the most likely description of one or more colors, the color that resulted in the highest probability of producing that term.

The results are in \figref{fig:wcs}. The charts use the layout of
the WCS stimulus, in which the two axes represent dimensions of color variation
similar to hue and lightness. Each region represents a set of colors that the model labeled with the same color term, and a star marks the color that resulted in the highest probability of producing that term. The Chinese terms, except for \textzh{红}{h\'ong}, are abbreviated by deleting the
final morpheme \textzh{色}{s\`e} `color'.

The charts agree with \citet{BerlinKay1969} on most of the differences between the two languages:
\emph{orange} and \emph{pink} have clear regions of dominance in English, whereas in the Mandarin
monolingual model \emph{pink} is subsumed by \textzh{红}{h\'ong} `red', and \emph{orange} is subsumed by \textzh{黄色}{hu\'angs\`e} `yellow'. Our models produce three colors not in the six-color system\footnote{Notably absent are `black' and `white'. The collection methodology of \citet{Monroe2017} restricted colors to a single lightness, so black and white are not in the data. For these charts, we replaced the World Color Survey swatches with the closest color used in our data collection.} identified by \citeauthor{BerlinKay1969} for Mandarin: \textzh{灰色}{hu\={\i}s\`e} `grey', \textzh{紫色}{z\v{\i}s\`e} `purple', and \textzh{棕色}{z\=ongs\`e} `brown'. We do not specifically claim these should be considered \emph{basic} color terms, since \citeauthor{BerlinKay1969} give a theoretical definition of ``basic color term'' that is not rigorously captured by our model. In particular, they explicitly exclude \textzh{灰色}{hu\={\i}s\`e} from the set of basic color terms, despite its frequency, because it has a meaning that refers to an object (`ashes'). The other two may have been excluded for the same reason, or they may represent a change in the language or the influence of English on the participants' usage.\footnote{MTurk's restriction to US workers makes English influence more likely than would otherwise be expected.}

A few differences between the monolingual and bilingual models can be characterized as an influence of
one language's color system on the other. First, \emph{teal} appears as a common description of a few color 
swatches from the English monolingual model, but the bilingual model, like the Chinese model, does not feature 
a common word for teal. Second, the Chinese monolingual model does not include a common word for orange, but
the bilingual model identifies \textzh{橙色}{ch\'engs\`e} `orange'. Finally, the English 
\emph{green} is semantically narrower than the Chinese \textzh{绿色}{l\textipa{\`{\"u}}s\`e}, and the
Chinese bilingual model exhibits a corresponding narrowing of the range of 
\textzh{绿色}{l\textipa{\`{\"u}}s\`e}.

Overall, however, the monolingual models capture largely accurate maps of each language's
basic color system, and the bilingual model retains the major contrasts between them, 
rather than ``averaging'' between the two. This suggests that
the bilingual model learns a representation of the input colors that
encodes their categorization in both languages, and that for the most part these
lexical semantic representations do not influence each other.

\subsection{Comparing model and human utterances}

\begin{figure}[!t]
    \begin{subfigure}[b]{\columnwidth}
    \centering
    \includegraphics[width=0.8\textwidth]{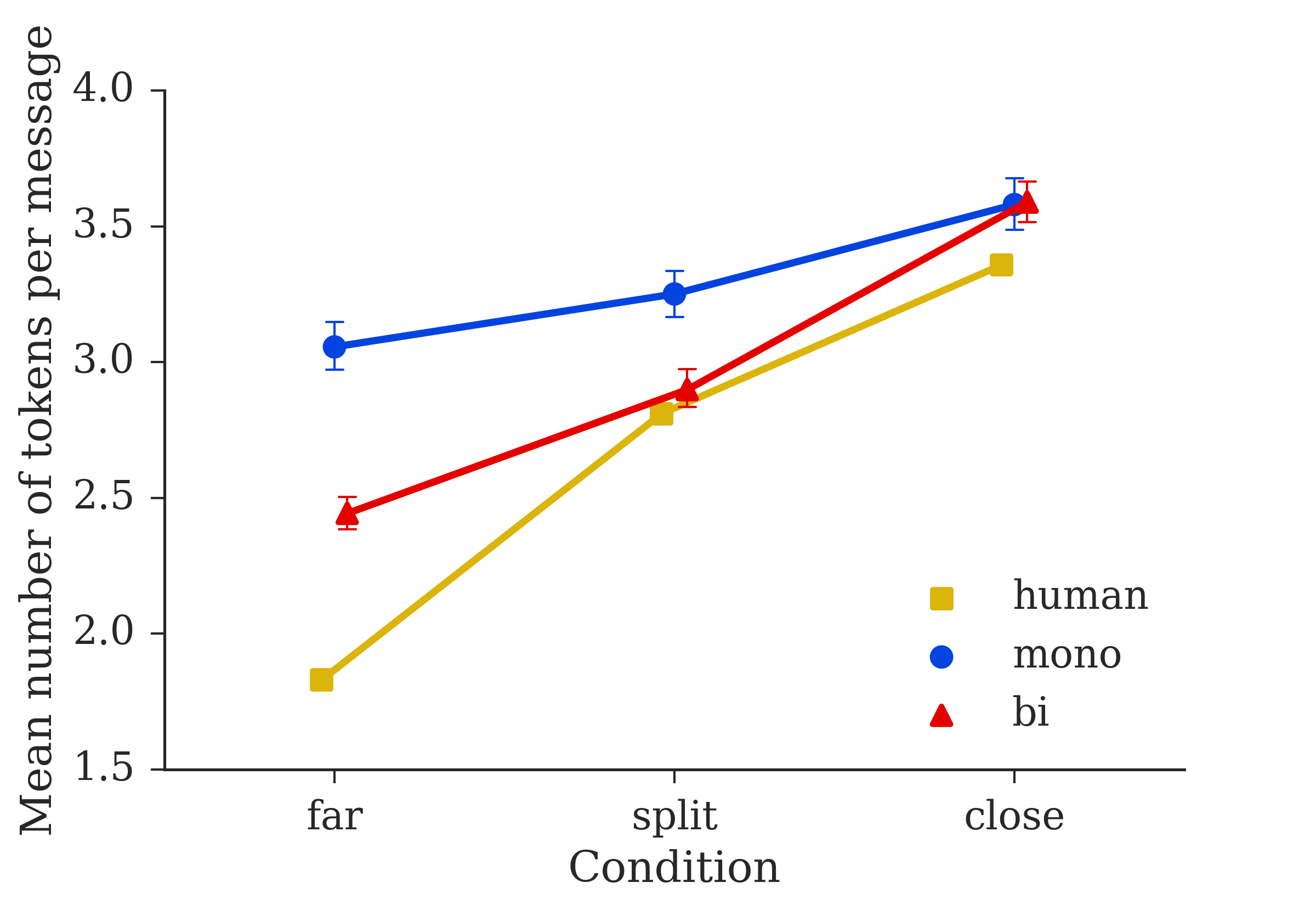}
    \caption{Human and model utterance lengths in English.}
    \label{fig:models_length:en}
    \end{subfigure}

    \begin{subfigure}[b]{\columnwidth}
    \centering
    \includegraphics[width=0.8\textwidth]{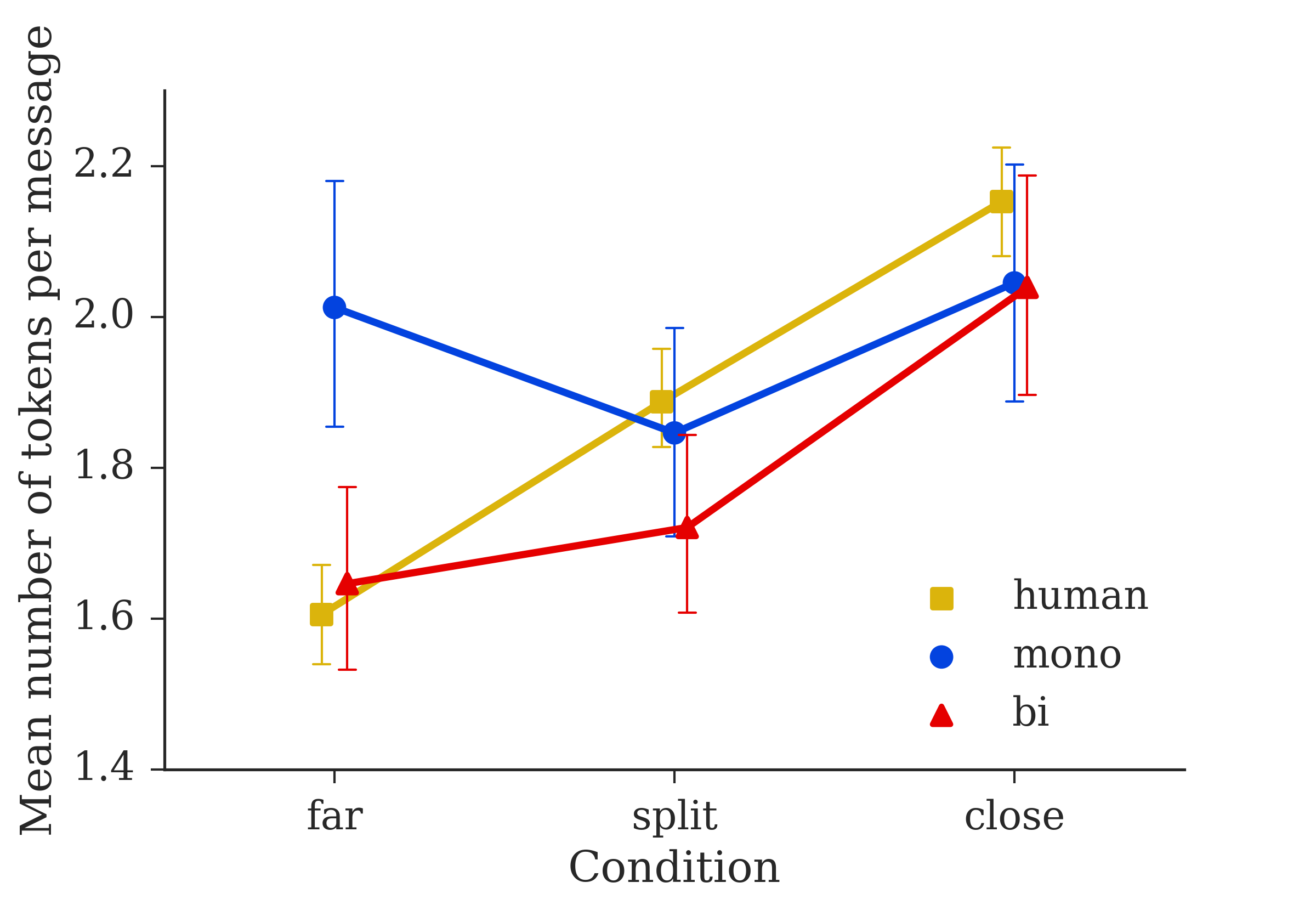}
    \caption{Human and model utterance lengths in Chinese.}
    \label{fig:models_length:zh}
    \end{subfigure}
\caption{Comparison of mean length of messages between human and model utterances.}
\label{fig:models_length}
\end{figure}

One observation indicates that the improvements in the bilingually-trained model are primarily at the pragmatic (context-dependent) level of language production. \figref{fig:models_length} reveals that the bilingually-trained model better
captures the main pragmatic pattern we observe in the human data, that of increasing message length in harder conditions. In both languages, the monolingual model
uses longer utterances in the easy \emph{far} condition than human speakers do, whereas the bilingual model is
significantly closer on that condition to the human statistics. We see similar results in the use of negations
and comparatives; the use of superlatives is not substantially different between the monolingual and bilingual
models.

We note that this result does not rule out several competing hypotheses. In particular,
we do not exclude improvements in compositional semantics or syntax, nor do we
distinguish improvements in specific linguistic areas from broader regularization
effects of having additional data in general. Preliminary experiments involving
augmentation of the data by duplicating and deleting constituents show no gains,
suggesting that the improvement depends on certain kinds of regularities in the English
data that are not provided by artificial manipulations. However, more investigation is
needed to thoroughly assess the role of general-purpose regularization in our
observations.

\section{Related work}

The method we use to build a bilingual model involves adding a single dimension to the previous-token vectors in the encoder representing the language (\secref{sec:models}). 
In essence, the two languages have separate vocabulary representation at the input and output but shared hidden representations. 
Adding a hard constraint on the output vocabulary
would make this equivalent to a simple form of multitask learning \citep{Caruana1997,Collobert2008}.
However, allowing the model to use tokens from either language at any time is simpler
and results in better modeling of mixed-language data, which is more common in
non-English environments. In fact, our model occasionally ignores the flag
and ``code-switches'' between the two languages within a single output, which is not possible in typical multitask architectures.

Using shared parameters for cross-lingual representation transfer has a large literature.
\citet{Klementiev2012} and \citet{Hermann2014} use multitask learning with multilingual document classification to build cross-lingual
word vectors, and observe accurate lexical translations from linear vector analogy operations. They include predicting translations for
words in parallel data as one of their tasks. Our translations from vector relationships (\secref{sec:lexicon-induction}) derive their cross-lingual
relationships from the non-linguistic input of our grounded task, without parallel data.

\citet{Huang2013} note gains in speech recognition from cross-lingual learning with shared parameters. In machine translation,
\citet{Johnson2016} add the approach of setting the output language using a symbol in the input. 
\citet{Kaiser2017} extend this to image captioning, speech recognition, and parsing in one multitask system.
Our work complements these efforts with an in-depth analysis of bilingual training on a 
grounded generation task and an exploration of the relationship between cross-lingual semantic differences 
and pragmatics. In general, we see grounding in non-linguistic input, including
images and sensory input from real and simulated worlds, as an intriguing
substitute for direct linguistic supervision in low-resource settings. We encourage
evaluation of multitask and multilingual models on tasks that require reference to 
the context for effective language production and understanding.

\section{Conclusion}

In this paper, we studied the effects of training on bilingual data in a grounded language task. We show evidence that
bilingual training can be helpful, but with a non-obvious effect of dataset size: accuracy as a function of opposite-language
data follows a U-shaped curve. The resulting model is more human-like in measures of sensitivity to contextual difficulty (pragmatics), while
exhibiting language-specific lexical learning in the form of vector relationships between lexical pairs
and differences between the two languages in common color-term extensions
(semantics).

It should be noted that color descriptions in English and Chinese are similar both in their syntax
and in the way they divide up the semantic space. We might expect that for languages like Arabic and Spanish (with their
different placement of modifiers), or Waorani and Pirah\~a (with their much smaller color term
inventories), the introduction of English data could have detrimental effects that 
outweigh the language-general gains. An investigation across a broader range of languages is desirable.

Our contribution includes a new dataset of human utterances in a color reference game in Mandarin Chinese, which we
release to the public\footnote{\url{https://cocolab.stanford.edu/datasets/colors.html}} with our code and trained model
parameters.\footnote{\url{https://github.com/futurulus/colors-in-context}}

\section*{Acknowledgments}

We thank Jiwei Li for extensive editing of our Chinese translations of the Mechanical Turk task instructions, Robert X.D.\ Hawkins for assistance setting up the data collection platform, and members of the Stanford NLP group---particularly Reid Pryzant, Sebastian Schuster, and Reuben Cohn-Gordon---for valuable feedback on earlier drafts. 
This material is based in part upon work supported by the Stanford Data Science Initiative and by the NSF under Grant Nos.~BCS-1456077 and SMA-1659585.

\bibliography{multilingual-naacl2018}
\bibliographystyle{acl_natbib}

\appendix

\section{Model details} \label{sec:appendix}

Hyperparameters for the two main models are given in \tabref{tab:hyperparameters}.
Both Chinese monolingual and bilingual model were tuned for perplexity on a held-out subset of the training set,
by random search followed by a local search from the best candidate until no single parameter change produced a better
result. However, the tuned settings for the Chinese monolingual model did not outperform the settings from
\citet{Monroe2017} for the English model on the development set, so in our final experiments the monolingual models
used the same parameters.

The vocabulary for each model consisted of all tokens that were seen at least twice in training; the bilingual
model's vocabulary is larger than the union of the words in each monolingual model because some tokens occurred once in each language
(largely meta-commentary---e.g., \word{dunno}, \word{HIT}, \word{xD}---and some English color word typos).

\begin{table}[!h]
\centering
\begin{tabular}{lccc}
\toprule
hyperparameter      & mono. & biling. \\
\midrule
optimizer           & ADAM  & RMSProp \\
learning rate       & 0.004 & 0.004 \\
dropout             & 0.1   & 0.1 \\
gradient clip norm  & --    & 1 \\
LSTM cell size      & 100   & 50 \\
embedding size      & 100   & 100 \\
initial forget bias & 0     & 5 \\
nonlinearity        & tanh  & sigmoid \\
\midrule
vocabulary size     & 895 (en) & 1,326 \\
                    & 260 (zh) & \\
\bottomrule
\end{tabular}
\caption{Values of hyperparameters optimized in tuning for the monolingual and bilingual models, plus vocabulary sizes.}
\label{tab:hyperparameters}
\end{table}

\section{Data management} \label{sec:irb}

This work was done under
Stanford IRB Protocol 17827, which has the title ``Pragmatic enrichment and contextual inference''.
Data was only collected from workers who indicated their informed consent.
Workers on Amazon Mechanical Turk were paid \$2.00 to complete each game consisting of 50 dialogue contexts, plus
a bonus of \$0.01 for each target the listener correctly identified.
All worker identifiers have been removed from data that is released; the only other information collected
about the workers was their Chinese language proficiency.

\end{document}